\documentclass[lettersize,journal]{IEEEtran}
\usepackage{amsmath,amsfonts,amssymb}
\usepackage{algorithmic}
\usepackage{algorithm}
\usepackage{array}
\usepackage[caption=false,font=normalsize,labelfont=sf,textfont=sf]{subfig}
\usepackage{textcomp}
\usepackage{stfloats}
\usepackage{url}
\usepackage{verbatim}
\usepackage{graphicx}
\usepackage{cite}

\usepackage{multirow}
\usepackage{booktabs}
\usepackage{color} 
\begin{document}

\title{PointGame: Geometrically and Adaptively Masked Auto-Encoder on Point Clouds}

\author{Yun Liu, Xuefeng Yan, Zhilei Chen, Zhiqi Li, Zeyong Wei, and Mingqiang Wei,~\textit{Senior Member, IEEE}

\thanks{Yun Liu, Xuefeng Yan, Zhilei Chen, and Zeyong Wei are with the School of Computer Science and Technology, Nanjing University of Aeronautics and Astronautics, Nanjing, China (e-mail: yun.liu.lydia@gmail.com, yxf@nuaa.edu.cn, zhil\_chen@outlook.com, weizeyong1@gmail.com). }

\thanks{Zhiqi Li is with the National Center for Computer Animation, Bournemouth University and the Alan Turing Institute, UK (e-mail: nanjzq16@gmail.com). }

\thanks{Mingqiang Wei is with the Shenzhen Research Institute, Nanjing University of Aeronautics and Astronautics, Shenzhen, China (e-mail: mingqiang.wei@gmail.com). }

}
\markboth{Journal of \LaTeX\ Class Files,~Vol.~14, No.~8, August~2021}%
{Shell \MakeLowercase{\textit{et al.}}: A Sample Article Using IEEEtran.cls for IEEE Journals}


\maketitle

\begin{abstract}
Self-supervised learning is attracting large attention in point cloud understanding. However, exploring discriminative and transferable features still remains challenging due to their nature of irregularity and sparsity. We propose a geometrically and adaptively masked auto-encoder for self-supervised learning on point clouds, termed \textit{PointGame}. PointGame contains two core components: GATE and EAT. GATE stands for the geometrical and adaptive token embedding module; it not only absorbs the conventional wisdom of geometric descriptors that captures the surface shape effectively, but also exploits adaptive saliency to focus on the salient part of a point cloud. EAT stands for the external attention-based Transformer encoder with linear computational complexity, which increases the efficiency of the whole pipeline. Unlike cutting-edge unsupervised learning models, PointGame leverages geometric descriptors to perceive surface shapes and adaptively mines discriminative features from training data. PointGame showcases clear advantages over its competitors on various downstream tasks under both global and local fine-tuning strategies. 
The code and pre-trained models will be publicly available.
\end{abstract}


\section{Introduction} 
\IEEEPARstart{A}{dvanced} 3D sensors, e.g., optical and laser scanners, are booming to digitize the real world~\cite{pami/ZhouCZW0WL0023}. The captured data from these sensors are originally represented by point clouds~\cite{tvcg/WeiCZXGW23}. Currently, point clouds are a mainstream yet very simple data representation form, serving various practical 3D applications such as self-automation driving, Metaverse, SLAM and robotics \cite{mingqiang-pami}. Recent years have witnessed many efforts of using deep learning to understand point clouds~\cite{wu2022casa, gulipeng}. Representation learning is often the first step for point cloud understanding~\cite{li2022global, huang20223, lixin}, which refers to extracting discriminative features from point clouds.
% 新增三篇TGRS的论文引用，一篇 点云object detection，一篇Hyperspectral(超光谱)对比学习，另一篇 HR Remote Sensing图像自监督对比学习
\par Representation learning on point clouds is still not well-solved. Currently, there mainly exist two representation learning paradigms for point clouds, i.e., supervised learning and unsupervised learning. For supervised learning, one requires considerable annotations of point clouds for training. However, unlike 2D images, point clouds consist of a series of disordered and irregular 3D points, which leads to the hard and inaccurate human annotations. This explains why we have few labeled point clouds of real-world scenarios nowadays. Although one can create 3D point cloud models and automatically label them with various geometric modeling tools, training on such `fake' data will inevitably suffer from domain shifts in real-scanned point clouds. In contrast, unsupervised learning can alleviate the shortage of labeled real-world data by absorbing the knowledge of point clouds themselves; it encourages to use massive unlabeled data and is invulnerable when applied in different scenes.  
\par As a popular unsupervised learning paradigm, self-supervised representation learning (SSRL) of point clouds aims at learning one hidden part of the input point cloud from another unhidden part. According to different pretext tasks, existing SSRL methods are categorized into contrastive and generative techniques. 
Contrastive methods impose different transformations on the given point cloud, and seek to perform feature space similarity measurement between positive and negative samples. 
By contrast, generative methods first encode the point cloud into the latent feature space, then randomly discard some feature tokens and restore the original 3D geometric structure from them. 

\par Although current SSRL methods~\cite{yu2022pointBERT, pang2022masked, zhang2022masked} have achieved impressive progresses on point clouds, they can be further improved from two aspects: 
(i) Multiple types of feature representation may be explored, which complement to each others to better perceive the global and local features of point clouds.  
(ii) It is of critical importance to distinguish different contributions of features in the channel and spatial axes, which is little discussed by current literature.

% ****************** 3.observation or assumption, 基于方法看到的和猜想的解决方案 ******************
\par To deal with the above-mentioned problems, we propose a geometrically and adaptively masked auto-encoder, dubbed PointGame, for self-supervised point cloud representation learning. 
Specifically, we first extract local geometry descriptors for point patches, and then design a geometrical and adaptive token embedding~(GATE) module to fuse patch features and geometry features, which can enrich the feature representation for better point cloud understanding. Since the attention mechanism is proved to be effective for representation learning in previous works \cite{mnih2014recurrent, xu2015show, jaderberg2015spatial}, we introduce an adaptive saliency module in GATE to highlight the beneficial features and suppress the unnecessary ones, thus improving the power of feature representation. The adaptive saliency module consists of a channel attention and a spatial attention to adaptively distinguish the informative features in both the channel and spatial axes. Given the geometry-aware robust feature tokens, we randomly mask some tokens as Point-MAE~\cite{pang2022masked}, and devise an external attention-based Transformer (EAT) encoder to generate deep features for subsequent reconstruction decoder. It is worth noting that we utilize external attention~\cite{guo2022beyond} instead of self attention~\cite{vaswani2017attention} to combine the correlations between different samples and reduce quadratic computation complexity. 
Comprehensive experiments demonstrate that PointGame achieves competitive performance with the state-of-the-art methods on both synthetic and real public benchmark datasets.

% ****************** 4.what we proposed and main contributions, 提出的点简述和贡献，一一对应 ******************
Our main contributions can be summarized as follows:
\begin{itemize}
\item We propose an efficient geometrically and adaptively masked auto-encoder for self-supervised learning on point clouds (PointGame). Unlike existing unsupervised learning models, PointGame leverages geometric descriptors to perceive surface shapes and adaptively mines discriminative features from training data.
\item We design a geometrical and adaptive token embedding module to fuse patch features and geometry features, which not only embeds the distinctive point cloud structure into feature representation, but also highlights the vital features in both the channel and spatial axes.
\item We introduce an external attention based Transformer encoder to generate deep features from unmasked feature tokens, which can takes into account both the correlations between different samples and computational complexity.
\end{itemize}
\par The remainder of this paper is organized as follows: Section \ref{sec:relatedWork} introduces the related work. Section \ref{sec:Method}  gives the details of our method. Section \ref{sec:Experiments} shows the experiments on downstream tasks, followed by conclusion in Section \ref{sec:Conculsion}. 

\IEEEpubidadjcol

\section{Related Work} \label{sec:relatedWork}
\subsection{Point Cloud Representation Learning}
\par Unlike 2D images that are arranged on regular grids, 3D point clouds are irregularly distributed in 3D space. As a result, it is hard for current CNNs to extract robust features from point clouds. A multitude of methods have been proposed to address this issue. We begin with a brief review of previous works on point cloud representation learning for integration purpose.
\par \textbf{Projection-based methods}~\cite{lang2019pointpillars, kanezaki2018rotationnet, su2015multi} tackle the unordered problem of point clouds by projecting them onto image planes. The resulting 2D images are then processed using 2D CNNs to extract point cloud features. Finally, the extracted features from each image are fused to form the final output features.
\textbf{Voxel-based methods}~\cite{maturana2015voxnet, song2017semantic} address the unordered problem by transforming the point cloud into 3D voxels, followed by 3D convolutions to extract point cloud features. However, this type of methods will result in massive computation and memory costs due to the cubic growth in the number of voxels as a function of resolution.
\textbf{Point-based methods}~\cite{qi2017pointnet, qi2017pointnet++, zhao2021point} directly process point clouds as sets embedded in continuous space. PointNet~\cite{qi2017pointnet} is a pioneer network, which operates directly on unordered points and propagates information via pooling operators. PointNet++~\cite{qi2017pointnet++} is proposed as a hierarchical feature learning paradigm to recursively capture local geometric structures, based on the PointNet network. DGCNN~\cite{wang2019DGCNN}  connects the point cloud into local graphs and dynamically computes these graphs to extract geometric features.

% Transformer是point-based的子集，一个encoder-decoder的结构，
\par The transformer structure has flourished in natural language processing (NLP) and 2D computer vision~(CV) tasks for a long time. Inspired by the successful applications of transformers in NLP ~\cite{vaswani2017attention, zhao2020exploring} and image region~\cite{he2022masked, he2022swin}, a lot of 3D vision backbones have been proposed. Transformer-based methods in 3D computer vision tasks follow an encoder-decoder architecture that comprises stacked self-attention and point-wise, fully connected layers. The self-attention mechanism is particularly well-suited for processing 3D point clouds because it can naturally operate on sets of points with positional attributes. 
Zhao et al.\cite{zhao2021point} and Guo et al.\cite{guo2021pct} have developed point transformer layers that utilize self-attention for various 3D scene understanding tasks, such as semantic segmentation, object part segmentation, and object classification. These methods with transformer architecture have salient benefits such as massively parallel computing, long-distance characteristics, and minimal inductive bias.

% 引入自监督
\par As is well-known, supervised representation learning methods rely heavily on large amount of labeled datasets and typically designed to address specific tasks and datasets, with limited transferability to other domains. To address the challenge of relying on large-scale annotated point cloud data, more researches focus on unsupervised representation learning with unlabeled point clouds, utilizing various paradigms of self-supervised learning pipelines. SSRL methods aim to produce general and robust features of unlabeled datasets which can transfer effectively to other datasets.

\subsection{Self-Supervised Representation Learning}
\par SSRL aims to acquire robust and general features from unlabelled data, thereby alleviating the burden of time-consuming data annotation and achieving superior performance in transfer learning. This is accomplished through the use of representation learning networks and well-designed pretext tasks. 
Recently, SSRL for point clouds has shown promising results in various shape analysis and scene understanding tasks. Generally, existing methods can be categorized into contrastive methods and generative methods.

% 讲对比学习的有哪些，对比学习本身存在什么问题，从而引出为什么我用生成式。
\par \textbf{Contrastive methods}~(e.g., STRL~\cite{huang2021STRL}, CrossPoint~\cite{afham2022crosspoint}) aim to train a network to learn discriminative features by distinguishing between positive and negative samples. 
Recent studies such as STRL~\cite{huang2021STRL} and CrossPoint~\cite{afham2022crosspoint} propose contrastive learning without negative pairs. These methods leverage point cloud transformations for augmentation and employ PointNet~\cite{qi2017pointnet} and DGCNN~\cite{wang2019DGCNN} as feature encoder. The loss function is constructed by minimizing the distance between augmented samples of the same object and maximizing the difference between samples of different objects.

\par \textbf{Generative methods} (e.g., OcCo~\cite{wang2021OCCO}, Point-BERT~\cite{yu2022pointBERT}, Point-MAE~\cite{pang2022masked}), employ a destroy-rebuild structure, wherein the encoder is used for effective feature abstraction and the decoder for reconstructing the original geometric structures.
For instance, the pre-text task of OcCo~\cite{wang2021OCCO} method is reconstructing the occluded points in a camera view. OcCo utilizes encoders such as PointNet~\cite{qi2017pointnet} and DGCNN~\cite{wang2019DGCNN} and applies the decoder to complete the occluded points. However, methods based on PointNet~\cite{qi2017pointnet} and DGCNN~\cite{wang2019DGCNN} require multiple pre-training for corresponding downstream tasks.
To address this issue, Point-MAE~\cite{pang2022masked} employs a mini-PointNet~\cite{qi2017pointnet} as the point embedding module to achieve permutation invariance. Similarily, MaskSurf~\cite{zhang2022masked} adds a normal prediction module to enhance point cloud understanding.
% 基于以上的研究，MAE框架适合 general and robust 表征的学习， 但是法向量不具有鲁棒性和描述力不强，因此需要更好的局部描述方法,因此引入描述子
Even though it has better performance than Point-MAE on real-world dataset, we argue that normal vectors are not sufficiently robust and descriptive to capture all the nuances in the data. Hence, we investigate geometric descriptors to extract robust, descriptive, and comprehensive local information through statistical analysis of neighboring point characteristics.

\par In contrast to methods that based on the standard transformer blocks with self-attention mechanisms, aligning with the objective of robust 3D representation learning, we employ an external attention block~\cite{guo2022beyond} to incorporate the correlations between different samples while reducing the computational complexity simultaneously.

\section{Proposed Method} \label{sec:Method}
\begin{figure*}[!t]
    \centering
    \includegraphics[width=1.0\linewidth]{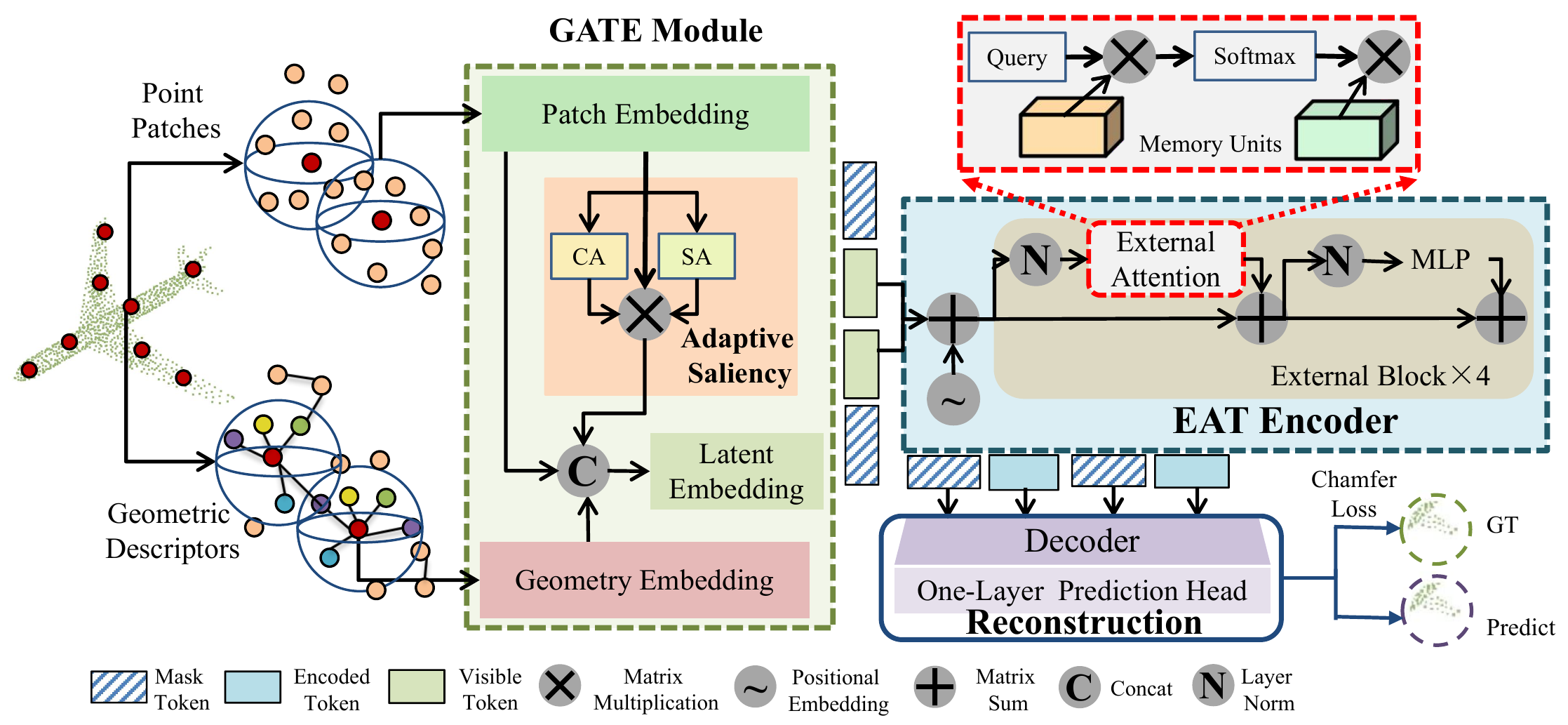}
    \caption{Framework of PointGame. Firstly, we propose geometrical and adaptive token embedding (GATE) module to generate latent tokens from point patches and geometric descriptors.
Next, our random mask strategy is applied to these latent tokens, followed by a transformer encoder with external attention to learn high-level latent features from unmasked tokens. 
Finally, we introduce a decoder module integrated with the self-attention layer and prediction head to reconstruct the masked point patches of the input point cloud. Please note that in the downstream tasks, the reconstruction module is avoided, and we only use the GATE and EAT modules as the backbone.}
    \label{fig:OverView}
\end{figure*}

\textbf{Overview.}  We propose a geometrically and adaptively masked auto-encoder, dubbed PointGame. PointGame seeks to  efficiently and effectively extract discriminative and transferable features in an unsupervised manner for downstream geometric tasks.
It mainly consists of a geometrical and adaptive token embedding (GATE) module, an external attention based Transformer (EAT) encoder, and a reconstruction module, as shown in  Fig.~\ref{fig:OverView}.

Firstly, we randomly select $g$ points from the whole point cloud, and construct a patch set $G=\{P_{i}|i=1,...,g\}$ by searching their local neighbor patches. For each of these selected $g$ points, we also excavate geometric descriptors to explore the geometry structure of the point cloud. 
Then, the point patches and geometric descriptors are fed into the GATE to obtain sufficiently robust and distinctive token embedding. Following the principle of MAE~\cite{he2022masked}, we randomly mask out some tokens, and utilize the EAT encoder to extract high-level latent features. Finally, the reconstruction module consumes the encoded tokens and initialized masked tokens, and predicts the masked point patches of the input point cloud. 

% 从局部和全局的角度来说，描述子获取局部细节信息，那么transformer能够获取全局信息？
\subsection{Geometrical and Adaptive Token Embedding}
\label{sec:GATE}
\begin{figure}[!t]
    \centering
    \includegraphics[width=1.0\linewidth]{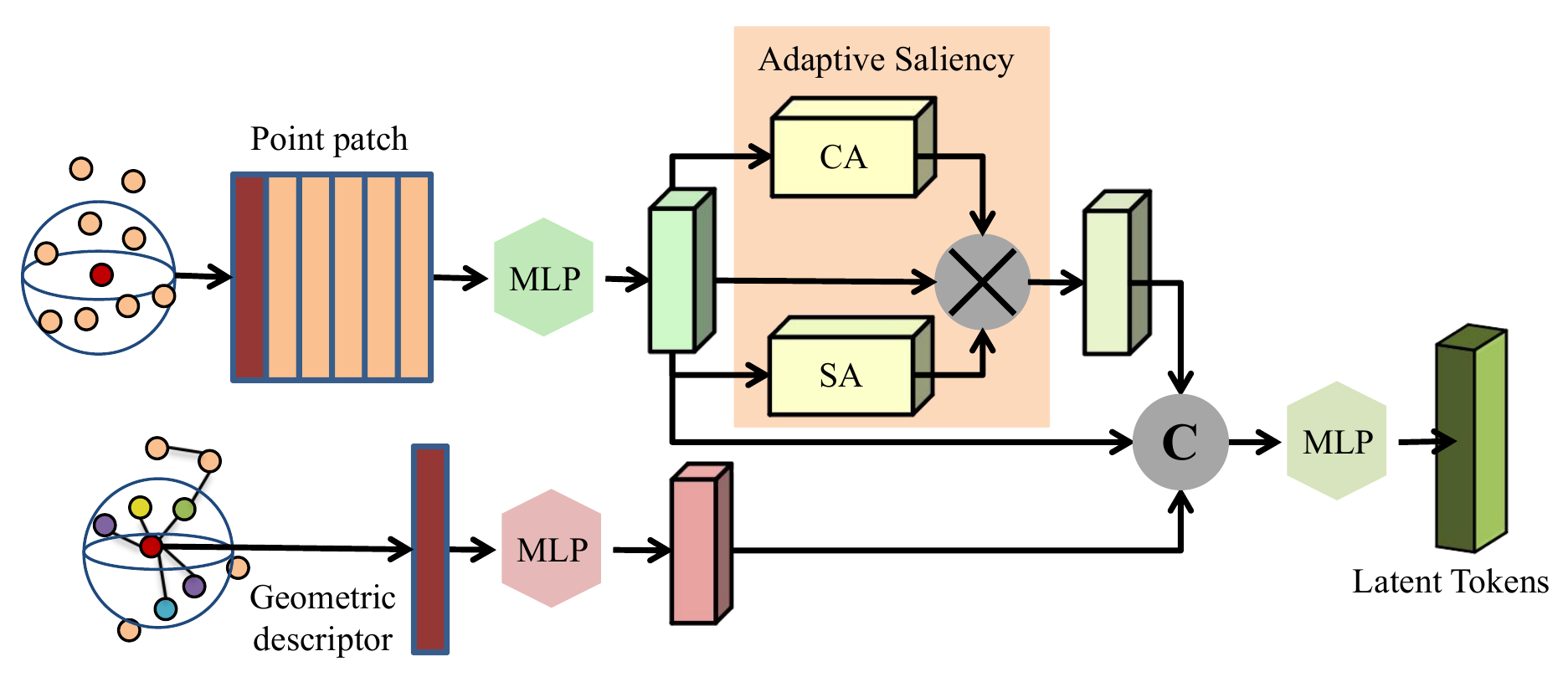}
    \caption{The GATE module. First, we construct the point patches and the geometric descriptors. Second, we feed each point patch and the descriptor of each patch's center point into individual MLP modules for patch embedding and geometry embedding, respectively. Third, the patch embedding is refined by an adaptive saliency attention module. Finally, all feature maps are concatenated and fed to another MLP module to obtain the latent tokens.}
    \label{fig:GATE}
\end{figure}

Accurate expression for a 3D shape is of prominent importance in point cloud understanding tasks. Previous works \cite{yu2022pointBERT, pang2022masked} adopt mini-PointNet~\cite{qi2017pointnet} to extract feature tokens for point patches, which ignore the local geometry structures of point clouds, and lack the descriptive and robust feature representation. 
This key observation inspires our geometrical and adaptive token embedding module (Fig. \ref{fig:GATE}), where geometric descriptors are integrated with point patches to represent 3D object shapes. 
For geometric descriptors, we utilize simplified point feature histograms (SPFH)~\cite{rusu2009fast} to perceive the details of nearby challenging feature regions, which can represent the underlying surface properties of a point. 
%Formally, we use 3D coordinates and estimated surface normals to compute SPFH descriptors. SPFH descriptors contain three histogram vectors of $\alpha$, $\phi$ and $\theta$, which can be expressed as, 
% Formally, SPFH descriptors consist of three histogram vectors of $\alpha$, $\phi$ and $\theta$. $\alpha$, $\phi$ and $\theta$ are computed by utilizing the 3D coordinates and estimated surface normals, which can be expressed as,  
Formally, we use 3D coordinates and estimated surface normals to compute SPFH descriptors, which can be expressed as follows:  

 % SPFH 计算方法和公式如下
\begin{equation*}
\alpha = {v}\cdot{n_q},
\label{eq:SPFH1}
\end{equation*}
\begin{equation}
\phi=\frac{({u}\cdot{(p_i-p_q)})} {||p_i-p_q||},
\label{eq:SPFH2}
\end{equation}
\begin{equation*}
\theta=arctan({w}\cdot{n_i},{u}\cdot{n_q}),
\label{eq:SPFH3}
\end{equation*}
where $p_q$ and $p_i$ represent the query point and its neighbor point, respectively. $n_q$ and $n_i$ denote the normal vectors of $p_q$ and $p_i$, respectively. $u$, $v$ and $w$ are defined as $u=n_q$, $v=(p_i-p_q)\times{u}$, $w={u}\times{v}$, forming the angular variations $\alpha$, $\phi$ and $\theta$.
The signature result $A_{pi}$ of the pair of $p_q$ and $p_i$ is the combination of histogram equalization statistics on $\alpha$, $\phi$ and $\theta$.
The geometric descriptor of the point $p_q$ is define as $D_{pq}=array(A_{pi}), i\in({0,...,k-1})$.
% 以上步骤是一个点p的描述子的计算过程
%However, this method is limited in extracting local geometric features based on coordinates. 
% To enable effective feature learning, we adopt geometric descriptors: Simplified point feature histograms (SPFH), 
%extracting robust and descriptive local features, respect the useful information from multiple types of features.
% perceiving the details in/near challenging feature regions and exploiting the useful information from multiple types of features. Therefore they can represent the underlying surface model properties at a point.

% 详述下 adaptive saliency module 的处理，并参考对应 figure
\begin{figure}[!t]
    \centering
    \includegraphics[width=1.0\linewidth]{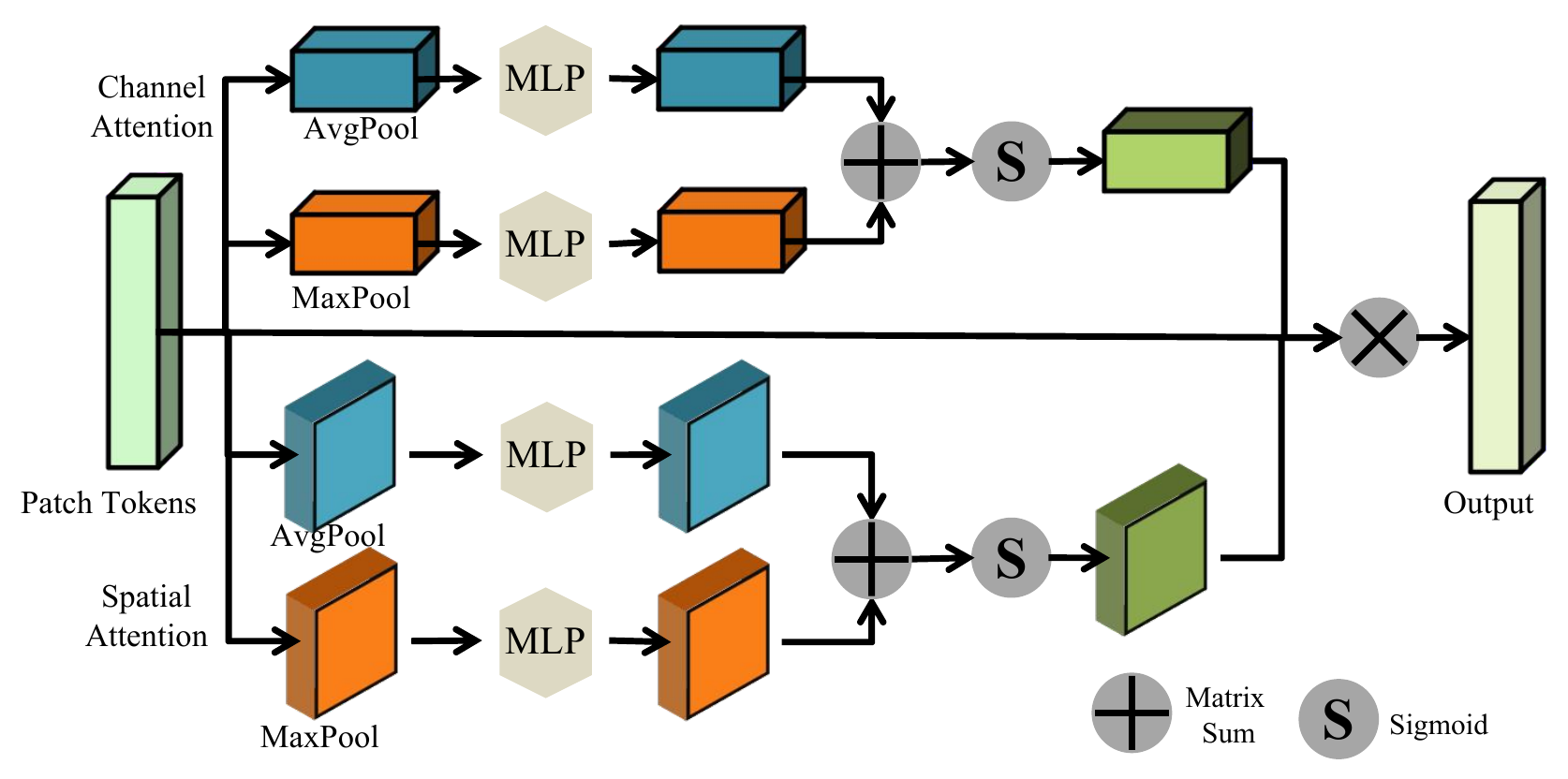}
    \caption{The adaptive saliency module.
    It accepts the patch tokens from the patch embedding module to obtain the salient features.}
    \label{fig:adaptiveSaliency}
\end{figure}

\par Since point clouds characterize 3D shapes, constructing point patches is necessary to investigate the geometrical cues within a local neighborhood. Given a point cloud, we first randomly sample $n$ points through Farthest Point Sampling~(FPS) and divide the sampled point set into $g$ irregular point patches~(may overlap) via  K-Nearest Neighborhood~(KNN) algorithm. Then, each point is normalized by subtracting the center point of its patch for better convergence. Note that we apply multi-layer perceptrons (MLPs) to embed geometric descriptors $D$ and point patches $G$ into high-dimensional features $P_T$ and $D_T$, respectively. Instead of fusing them directly, we attempt to pay more attention to the distinctive features contained within point patches. Therefore, we employ a Channel Attention~(CA) and a Spatial Attention~(SA) on $P_T$ in parallel, as shown in Fig.~\ref{fig:adaptiveSaliency}. This can be viewed as the process of selecting salient attributes from the input patch tokens. The calculation of salient feature tokens is as follows:
\begin{equation*}
\begin{split}
W_{CA}= Sigmoid(MLP(AvgPool_{CA}(P_T)) \\ +MLP(MaxPool_{CA}(P_T))),
\label{eq:CA}
\end{split}
\end{equation*}
\begin{equation}
\begin{split}
W_{SA}= Sigmoid(MLP(AvgPool_{SA}(P_T)) \\ +MLP(MaxPool_{SA}(P_T))),
\label{eq:SA}
\end{split}
\end{equation}
\begin{equation*}
S_T=P_T \times{W_{CA}} \times{W_{SA}},
\label{eq:saliency}
\end{equation*}
where $Sigmoid$ denotes the sigmoid activation function. $AvgPool_{x}, x \in \{CA\}$ is the average pooling operation applied in CA. $MaxPool_{x}, x \in \{SA\}$ represents the max pooling function in SA. Then the salient patch tokens $S_T$ are calculated by matrix multiplication between channel weights and spatial weights. 
GATE differs from the existing methods because it emphasizes vital features and suppresses unnecessary features, which allows a further improvement of feature representation capability of our model. Next, $P_T$, $S_T$ and $D_T$ are concatenated and processed by an MLP~(Latent Embedding module). A max pooling layer is leveraged to aggregate features of patches for the latent feature tokens $T\in\mathbb{R}^{g\times{d}}$:
\begin{equation*}
T=MaxPool(MLP(cat(P_T,S_T,D_T))),
\label{eq:globalToken}
\end{equation*}
where $cat(.,.)$ denotes the concatenation, and $d$ is the dimension of features.
% 3D shape的表示很重要，但是目前的的方法没有考虑到几何结构，并且特征表示也不鲁棒 -> 因此，我们引入了几何算子SPFH -> 除了几何算子，我们也构造了patch -> 首先对这两个特征分别做embedding -> 对于patch feature，用adaptive saliency ->最后将patch feature和geometry feature融合得到latent token
% 拿到CA-SA权重后，要 feature* weight_CA*weight_SA 
% \par We feed the patch tokens $P_T$ to adaptive saliency module, obtaining the parallel spatial position attention weight $W_{CA}$ and channel attention weight $W_{SA}$ for reallocating computation resources. \textcolor{blue}{Then the salient patch tokens $S_T$ are calculated by matrix multiplication between channel weights and spatial weights.} The formulas are as follows:
% 三个特征concat，再经过MLP得到 fusion 的特征，这个特征会先mask再给到EAT
% In the last step, we concatenate $P_T$, $S_T$ and $D_T$ as the input of the last MLP module~(Latent Embedding module). The feature dimension of these embedding modules is $d$. We get the global latent tokens via a max pooling layer, which is denoted as $T\in\mathbb{R}^{g\times{d}}$. 

\subsection{External Attention-based Transformer Encoder}
\begin{figure}[!t]
    \centering
    \includegraphics[width=1.0\linewidth]{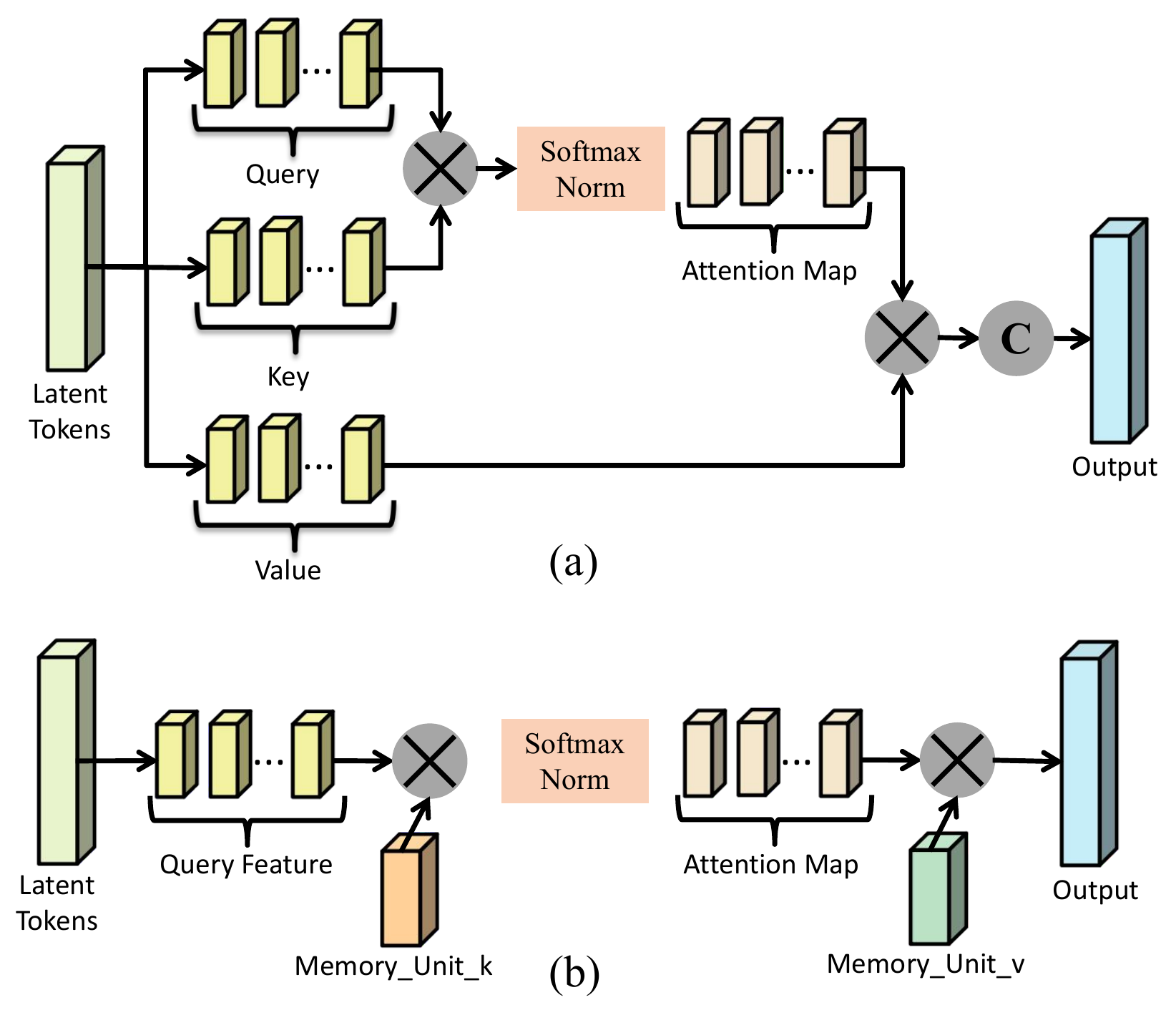}
    \caption{The EAT module. EAT utilizes the external attention instead of the self-attention in transformer blocks. EAT `eats' the visible latent tokens to obtain the high-level features with linear computational complexity. (a) self-attention mechanism and (b) external attention mechanism.}
    \label{fig:selfExternalAtt}
\end{figure}

% EAT是MAE的decoder部分的改进， 简述下我们方法的基于的MAE架构，实现承上启下：% 继承了MAE的思想，我们接下来mask tokens， encoder-decoder
Masked auto-encoding~\cite{he2022masked} masks a portion of input data and adopts an auto-encoder to reconstruct explicit features~(e.g., pixels) or implicit features~(e.g., discrete tokens) corresponding to the original masked content. Intuitively, the masked parts will not contribute any data information, hereby the auto-encoder could learn high-level latent features from unmasked parts. Following this idea, we first randomly mask tokens with a masking ratio $r$ to obtain the visible latent tokens $T^{v}$. Next, we  utilize an MLP to embed the center coordinates of visible patches into positional tokens $L_T^{v}$. We observe that most existing methods~\cite{pang2022masked, zhang2022masked, jiang2022masked} based on MAE leverage standard Transformer~\cite{vaswani2017attention} for self-supervised learning, which has a quadratic computational complexity and ignores the potential correlations between different data samples. Taking visible features tokens $T^{v}$ and positional tokens $L_T^{v}$ as inputs, we propose an external attention based Transformer encoder to excavate deep high-level latent features while minimizing the computational cost. The applied external attention block in Fig.~\ref{fig:selfExternalAtt}~(a) is compared with standard self-attention block in Fig.~\ref{fig:selfExternalAtt}~(b), and the definition of encoded tokens $T_E$ with external attention is as follows: 
%The external attention mechanism is first proposed in ~\cite{guo2022beyond} to fully explore correlations of samples with a much lower computational complexity. 
\begin{equation}
\begin{split}
T_E=Encoder(T_{v},L_T^{v}), \\ T_E\in\mathbb{R}^{(1-r)\times{g}\times{d}}.
\label{eq:encoder}
\end{split}
\end{equation}
% 插入Beyond external中的Figure 1 的a和c图，对比表示self-attention和external-attention，放到方法部分，我的方法中，encoder采用的是external attention， decoder采用的是self-attention
%\par \textcolor{blue}{In the preceding stage, we construct the inputs and embed for tokens with GATE module. }
% 经过GATE，得到fusion的tokens，这里mask tokens，这里有问题，我是得到mask这个蒙版向量(bz,64)，标记64个groups 和 tokens向量哪些是被mask掉的

\subsection{Reconstruction}
\par Following MAE, the pre-text task is the reconstruction of the masked points $P_{GT}\in\mathbb{R}^{r\times{g}\times{k}\times{3}}$. We employ an MLP for position embedding, which shares weights for masked and unmasked centers, $L_{TR}^{m}$ and $L_{TR}^{v}$.
Then, we initialize the masked token $T^{m}\in\mathbb{R}^{r\times{g}\times{d}}$ by duplicating a learnable masked token of $d$ dimension. The decoded tokens $T_D$ is formulated as, 
\begin{equation}
\begin{split}
T_D=Decoder(cat(T_E,T^{m}),cat(L_{TR}^{m},L_{TR}^{v})), \\  T_D\in\mathbb{R}^{r\times{g}\times{d}}.
\label{eq:decoder}
\end{split}
\end{equation}

\par Following Point-BERT~\cite{yu2022pointBERT} and Point-MAE~\cite{pang2022masked}, we adopt a fully connected~(FC) layer for rebuilding the masked point patches. We feed the decoded tokens $T_D$ and obtain the predicted masked point patches $P_{pre}$ as, 
\begin{equation}
\begin{split}
P_{pre}=Reshape(FC(T_D)), \\ P_{pre}\in\mathbb{R}^{r\times{g}\times{k}\times{3}}.
\label{eq:reconstruction}
\end{split}
\end{equation}

% \subsection{Loss function}
\par The loss function is based on the predicted masked point patches $P_{pre}$ and their ground-truth counterpart $P_{GT}$. We adopt $L_2$ Chamfer Distance~\cite{fan2017point} to compute the reconstruction loss as, 
% P_{GT}  and  P_{pre}
\begin{equation}
\begin{split}
L=\frac{1}{|P_{pre}|}\sum_{\hat{x}\in{P_{pre}}}\min_{x\in{P_{GT}}}||\hat{x}-x||_2^2 \\ +\frac{1}{|P_{GT}|}\sum_{x\in{P_{GT}}}\min_{\hat{x}\in{P_{pre}}}||\hat{x}-x||_2^2.
\label{eq:loss}
\end{split}
\end{equation}

\section{Experiment} \label{sec:Experiments}
\par In this section, we initially introduce the pre-training process of our model on ShapeNet55~\cite{chang2015shapenet} and demonstrate its performance on downstream tasks in Sec.~\ref{sec:pretrain}. Subsequently, we will evaluate the efficacy of our pre-trained model by utilizing shape classification tasks on ModelNet40~\cite{wu2015_modelNet} and ScanObjectNN~\cite{uy2019revisiting} in Sec.~\ref{sec:ShapeClassification}. The performance of our pre-trained model shall be evaluated on the task of part segmentation on ShapeNetPart~\cite{yi2016scalable} in Sec.~\ref{sec:PartSeg}. In Sec.~\ref{sec:SemSeg}, we will present the results of our experiments on semantic segmentation with S3DIS~\cite{armeni20163d}. Finally, we will conduct multiple ablation experiments to scrutinize and substantiate the efficacy of our designed modules in Sec.~\ref{sec:Ablation}.

\subsection{Pre-training on ShapeNet} \label{sec:pretrain}
\par Our method is pre-trained with the ShapeNet55~\cite{chang2015shapenet}, which contains 52,470 individual clean 3D models covering 55 commonly occurring object categories. The dataset is partitioned into a training set consisting of 41,952 instances and a validation set consisting of 10,518 instances. 

\par Our model is pre-trained solely on the training set, using fixed-size point clouds consisting of $n=1,024$ points as input data. Standard random scaling and translation techniques are applied for data augmentation. 
We utilize the FPS algorithm to sample $g=64$ points as group centers from the $1,024$ points, and then search $k=32$ points around every center point through KNN algorithm. 
We randomly mask the groups with a masking ratio of $r=60\%$. The encoder module is composed of $12$ transformer blocks based on external attention, while the decoder module comprises of $4$ standard transformer blocks based on self-attention. Each transformer block has $384$ hidden dimensions with $6$ heads, and the MLP ratio in transformer blocks is set to $4$. We adopt the AdamW optimizer~\cite{loshchilov2017decoupled} and cosine learning rate decay~\cite{loshchilov2016sgdr}. The initial learning rate is set to $0.001$ and the weight decay is $0.05$. Our model is pre-trained with $300$ epochs, with a batch size of $128$.

\par To perform a precise evaluation of our pre-trained model, we will employ two fine-tuning strategies for downstream tasks: global fine-tuning and local fine-tuning. Global fine-tuning involves fine-tuning all parameters, including the pre-trained encoder parameters and new parameters for downstream tasks. 
Local fine-tuning entails freezing all pre-trained parameters and only fine-tuning the parameters of downstream tasks. For shape classification, two categories are utilized: linear classification and non-linear classification, based on the type of classifier.

% % Note that the predicted coordinates of point patches must be compared with the normalized data. If we want to visualize the predicted results, the visualized coordinates are equal to the predicted coordinates plus the center points. 
% 参考PointBERT figure 2，列出ModelNet40：(input, masked input, output), ScanObject：(input, masked input, output)
\par
\begin{figure*}[!t]
    \centering
    \includegraphics[width=1.0\linewidth]{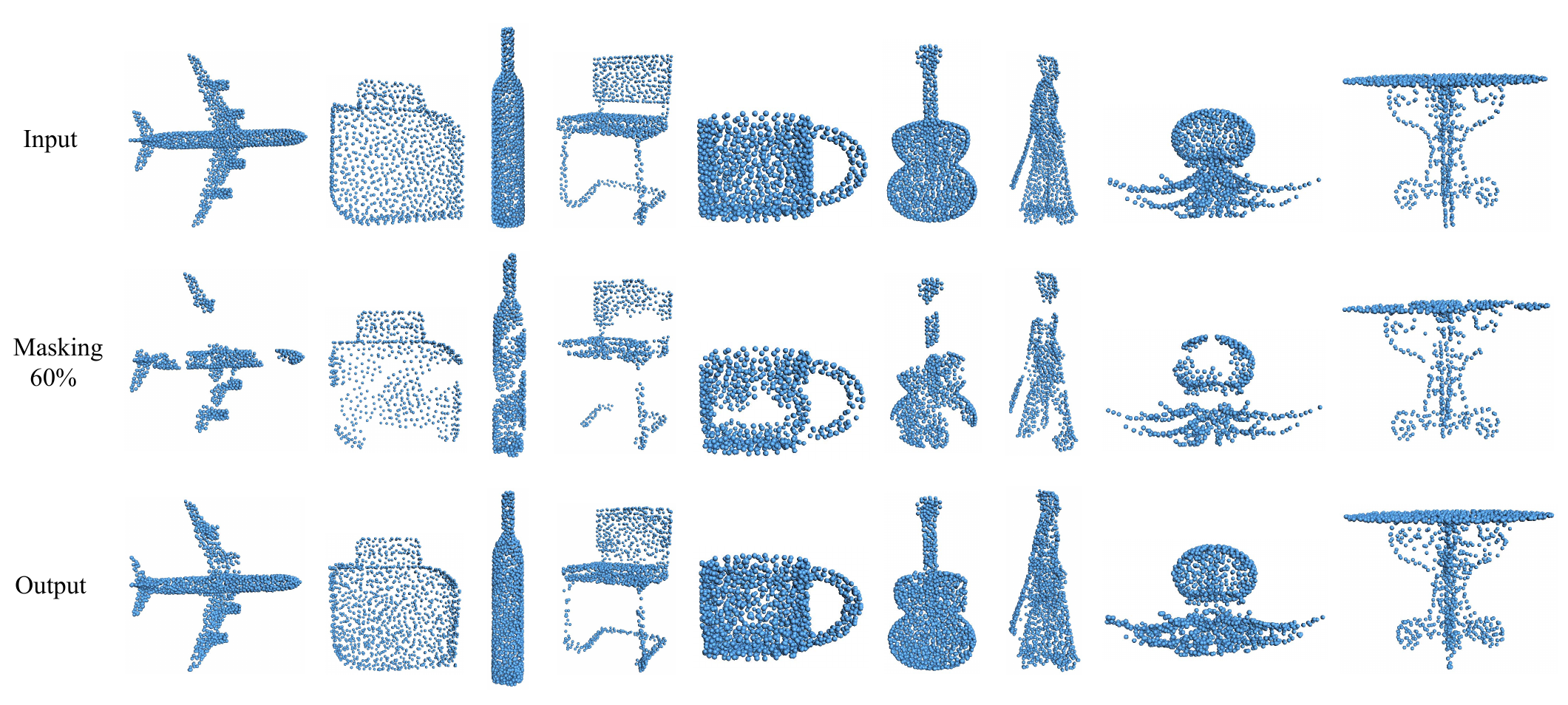}
    \caption{Visualization of raw input point clouds, visible patches masked 60\% and reconstruction results on ModelNet40.}
    \label{fig:reconstruction}
\end{figure*}

 We will examine the effect of initializing our self-supervised model with pre-trained backbone weights. The reconstruction results of our pre-trained model are demonstrated with ModelNet40~\cite{wu2015_modelNet}, as depicted in Fig.~\ref{fig:reconstruction}. 
This figure demonstrates that our pre-trained model achieves excellent reconstruction performance on the ModelNet dataset. Additionally, we observe that regions with high masking density lead to less satisfactory reconstruction outcomes compared to the regions that are normally sampled.
% 但是，如果一个物体的某个区域被集中mask掉，这块区域重建效果较差 

\subsection{Shape Classification} \label{sec:ShapeClassification}
\par We assess the performance of our pre-trained model through shape classification using the widely used ModelNet40~\cite{wu2015_modelNet} dataset, which consists of 12,311 clean 3D CAD models, covering 40 object categories, comprising of a training set with 9843 instances and a testing set with 2468 instances. 
To augment the data, we employ random scaling and random translation during experimentation. 
In the task of shape classification, we utilize two fine-tuning strategies, namely global fine-tuning and local fine-tuning, and evaluate their performance using both linear and non-linear classifiers.
For fair comparisons, we adopt the standard voting method~\cite{liu2019relation} during testing in this shape classification experiment on ModelNet40.

% $``\text{Pre-trained}"$, $``\text{Rep.}"$ 
\par The tables below exhibit the experimental outcomes, with `Pre-trained' denoting whether the model is supervised or unsupervised with a pre-trained model, and `Rep.' indicating that the results are reproduced by us using the official codes. Since the Point-MAE~\cite{pang2022masked} solely presents the outcomes for global fine-tuning in its official implementation, we conduct all shape classification experiments with both local and global fine-tuning techniques, utilizing the pre-trained model of Point-MAE available in the official codebase, and replicate all experimental settings from our previous work.

\par \textbf{Shape Classification on synthetic dataset.}
As presented in Tab.~\ref{tab:shapeClsMN40nonLinearGfinetune}, we employ a non-linear classifier with global fine-tuning on the ModelNet40 classification task. 
Our methodology results in an accuracy of 0.3\% higher than that of Point-MAE (93.05\%), and 0.15\% higher than the reproduced version of Point-BERT (93.2\%). Notably, when compared to other self-supervised learning approaches, our proposed PointGame outperforms all others in terms of performance.

%%%%%%%%%%%% 备注，基于作者给的pretrain模型训练分类：92.8，基于我自己预训练的模型训练分类：93.05%
\begin{table}[!t]
\caption{Object classification on ModelNet40 based on non-linear classifier with the global fine-tuning protocol.}
\centering
\begin{tabular}{c|cc}
\toprule  
Methods             & Pre-trained & Acc.  \\
\hline
PointNet~\cite{qi2017pointnet}    & -           & 89.2  \\
PointNet++~\cite{qi2017pointnet++}  & -           & 90.7  \\
PointCNN~\cite{li2018pointcnn}    & -           & 92.5  \\
DGCNN~\cite{wang2019DGCNN}       & -           & 92.9  \\
RS-CNN~\cite{liu2019relation}      & -           & 92.9  \\
PCT~\cite{guo2021pct}         & -           & 93.2  \\
PVT~\cite{zhang2021pvt}         & -           & 93.6  \\
PointTransformer~\cite{zhao2021point}         & -           & 93.7  \\
%Transformer~\cite{vaswani2017attention}         & -           & 91.4  \\
\hline
OcCo+DGCNN~\cite{wang2021OCCO}  & Y           & 93.0  \\
STRL+DGCNN~\cite{huang2021STRL}  & Y           & 93.1  \\
FoldingNet~\cite{yang2018foldingnet}  & Y           & 93.1  \\
Point-BERT~\cite{yu2022pointBERT}  & Y           & 93.2  \\
Point-MAE~\cite{pang2022masked}   & Y           & 93.8  \\
Point-MAE (Rep.) & Y           & 93.05 \\
Ours      & Y           & 93.35 \\
\bottomrule
\end{tabular}
\label{tab:shapeClsMN40nonLinearGfinetune}
\end{table}

\par Subsequently, we freeze the parameters of the pre-trained model, and solely fine-tune the classifier module. The classification results are presented in Tab.~\ref{tab:shapeClsMN40nonLinearLfinetune}. Our proposed methodology brings a boost of 0.16\% compared to Point-MAE~(Rep.).

\begin{table}[!t]
\caption{Object classification on ModelNet40 based on non-linear classifier with local fine-tuning protocol.}
\centering
\begin{tabular}{c|cc}
\toprule
Methods             & Pre-trained & Acc.  \\ \hline
Point-MAE (Rep.) & Y           & 92.91 \\ 
% Point-MAE (Rep.) & Y           & 92.66 \\
Ours      & Y           & 93.07 \\
% Ours      & Y           & 92.50 \\
\bottomrule
\end{tabular}
\label{tab:shapeClsMN40nonLinearLfinetune}
\end{table}

\par In Tab.~\ref{tab:shapeClsMN40LinearLfinetune}, we freeze the parameters of the pre-trained model and solely fine-tune the linear classifier module. Our proposed methodology achieves significantly superior performance (93.19\%) in comparison to Point-MAE (92.71\%) and IAE+DGCNN (92.1\%).

\begin{table}[!t]
\caption{Object classification on ModelNet40 based on linear classifier with the local fine-tuning protocol.}
\centering
\begin{tabular}{c|cc}%每列之间的|代表整列都要绘制|
\toprule
Methods             & Pre-trained & Acc.  \\ \hline
Multi-Task+DGCNN~\cite{hassani2019unsupervised}  & Y           & 89.1  \\
Self-Constrast+DGCNN~\cite{du2021self}  & Y           & 89.6  \\
Jigsaw+DGCNN~\cite{sauder2019self}  & Y           & 90.6  \\
FoldingNet+DGCNN~\cite{yang2018foldingnet}  & Y           & 90.1  \\ 
Rotation+DGCNN~\cite{poursaeed2020self}  & Y           & 90.8  \\

STRL+DGCNN~\cite{huang2021STRL}  & Y           & 90.9  \\
OcCo+DGCNN~\cite{wang2021OCCO}  & Y           & 89.2  \\

CrossPoint+DGCNN~\cite{afham2022crosspoint}  & Y           & 91.2  \\
IAE+DGCNN~\cite{yan2022implicit}  & Y           & 92.1  \\ 

Point-MAE (Rep.) & Y           & 92.71 \\ % acc=92.14
%Point-MAE(Rep.) & Y           & 92.14 \\
Ours      & Y           & 93.19 \\  %acc=92.34
%Ours      & Y           & 92.34 \\
\bottomrule
\end{tabular}
\label{tab:shapeClsMN40LinearLfinetune}
\end{table}

%%%%%%%%%%%%%%%%%%%%% t-SNE基于预训练模型得到concat_f768和non-linear的分类得到的特征 %%%%%%%%%%%%%%%%%%%%%
% \subsubsection{features of two datasets, PointBERT figure4, t-SNE[48]}
% 为了测试shapenet55预训练模型效果,我们可视化:点云经过pre-trained model后得到的特征向量的分布以及经过分类头后的分布
\par
\begin{figure}[!t]
    \centering
    \includegraphics[width=1.0\linewidth]{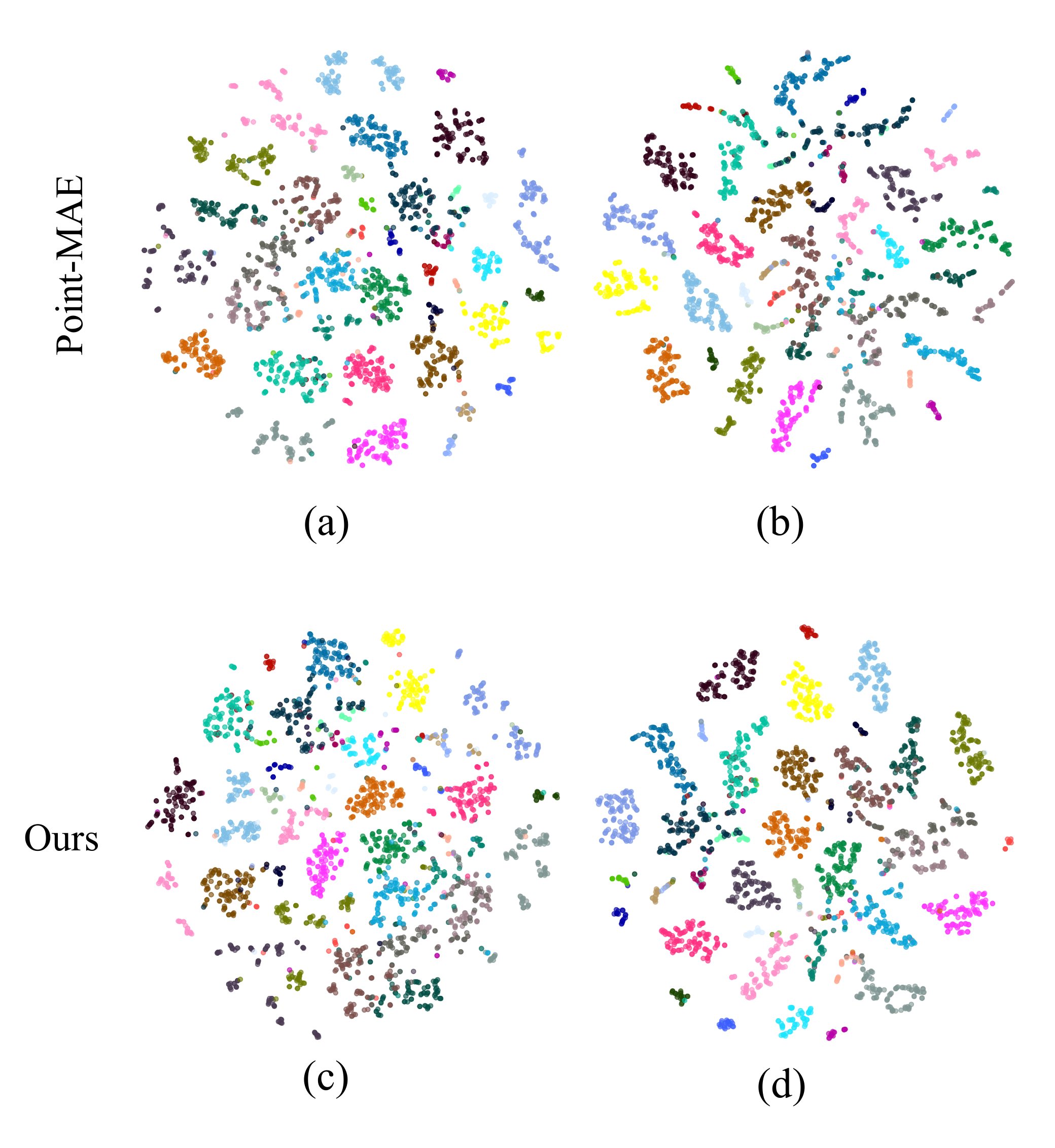}
    \caption{Visualization of feature distributions. We show the t-SNE visualization of feature vectors learned by Point-MAE~\cite{pang2022masked} and ours. (a) after pre-trained model and (b) after fine-tuning on ModelNet40 by PointMAE. (c) after pre-training, (d) after fine-tuning always on ModelNet40 by our method.}
    \label{fig:tSNEfeatures}
\end{figure}

 \textbf{Visualize the distribution of features.} In order to validate our pre-trained model based on ShapeNet55 and non-linear shape classifier on ModelNet40 with global fine-tuning protocol, and we employ t-SNE~\cite{van2008visualizing} to visualize the feature vectors of our pre-trained model and classifier in Fig.~\ref{fig:tSNEfeatures}. 
To facilitate a meaningful comparison, we opt for Point-MAE~\cite{pang2022masked} as our baseline and initially generate feature vectors by employing its pre-trained model and classifier, which are accessible in the official codebase. 
The dimensionality of the feature vectors derived from the pre-trained model $F_{pre}$ is $d=768$, while the result vectors from the non-linear classifier $F_{cls}$ have a dimensionality of $d=40$.
We employ t-SNE to depict the learned features $F_{pre}$ and $F_{cls}$, with the feature vectors of Point-MAE and our method. 
Fig.~\ref{fig:tSNEfeatures}~(a) and (b) show the feature vectors of Point-MAE's pre-trained module and classifier, respectively.
Subsequently, we utilize similar steps to calculate the feature vectors of our proposed approach and visualize them in Fig.\ref{fig:tSNEfeatures}~(c) and (d).

\par \textbf{Few-shot Learning.} Following previous works~\cite{sharma2020self, yu2022pointBERT, wang2021OCCO, pang2022masked, zhang2022masked}, we conduct few-shot learning experiments with pre-trained model on ModelNet40~\cite{wu2015_modelNet}. 
We employ an $n$-way, $m$-shot setup, where $n$ denotes the number of classes randomly selected from the dataset, and $m$ represents the number of objects randomly sampled for each class. We leverage ${n}\times{m}$ objects for training, and for evaluation purposes, we randomly select 20 objects from each of $n$ classes, which are unseen during training.
\par In the experiments, we set $n\in{5,10}$ and $m\in{10,20}$ and present the results in Tab.~\ref{tab:fewShotMN40Gfinetune}. Our method achieves comparable results to Point-MAE~\cite{pang2022masked}.

\par 
\begin{table}[!t]
\caption{Few-shot object classification on ModelNet40. We conduct 10 independent experiments for each setting and report mean accuracy (\%) with standard deviation.}
% \resizebox{8cm}{8mm}{  % \multirow{2}{*}{$mIoU_i$}
\centering
\begin{tabular}{c|cc}
\toprule
\multirow{2}{*}{Methods} & 5-way,10-shot & 5-way,20-shot\\ & 10-way,10-shot & 10-way,20-shot \\ \hline
\multirow{2}{*}{DGCNN-rand~\cite{wang2021OCCO}} & $31.6 \pm 2.8$  & $40.8 \pm 4.6$ \\& $19.9 \pm 2.1$    & $16.9 \pm1.5 $          \\ \hline
\multirow{2}{*}{DGCNN-OcCo~\cite{wang2021OCCO}} & $90.6 \pm 2.8$   & $92.5 \pm 1.9$ \\ & $82.9 \pm 1.3$   & $86.5 \pm 2.2$          \\ \hline
\multirow{2}{*}{Transformer-rand~\cite{yu2022pointBERT}} & $87.8 \pm 5.2$  & $93.3 \pm 4.3$  \\      & $84.6 \pm 5.5$         & $89.4 \pm 6.3$         \\ \hline
\multirow{2}{*}{Transformer-OcCo~\cite{yu2022pointBERT}} & $94.0 \pm 3.6$  & $95.9 \pm 2.3$  \\      & $89.4 \pm 5.1$         & $92.4 \pm 4.6 $        \\ \hline
\multirow{2}{*}{PointBERT~\cite{yu2022pointBERT}}  & $94.6 \pm 3.1$        & $96.3 \pm 2.7$ \\       & $91.0 \pm 5.4 $        & $92.7 \pm 5.1$         \\ \hline
\multirow{2}{*}{PointMAE~\cite{pang2022masked}}  & $96.3 \pm 2.5$        & $97.8 \pm 1.8$   \\     & $92.6 \pm 4.1$         & $95.0 \pm 3.0$         \\ \hline
\multirow{2}{*}{Ours}    & $96.1 \pm 3.1$         & $97.9 \pm 1.4$  \\   & $90.3 \pm 5.1$         & $93.8 \pm 2.5$          \\ 
\bottomrule
\end{tabular}%}
\label{tab:fewShotMN40Gfinetune}
\end{table}

\par \textbf{Shape Classification on the real-world dataset.} 
The objective of SSRL is to design a model with strong capability of domain adaptation, which trains on the synthetic datasets while still being able to perform well on real-world datasets.
Real scenes are often more intricate than synthetic scenes, and can involve multiple objects, some of which may be incompletely scanned, posing a significant challenge to shape classification. In this paper, we design a self-supervised learning approach with high domain adaptation capabilities to address this challenge.
We evaluate the performance of our pre-trained model on the ScanObjectNN~\cite{uy2019revisiting} dataset, which contains approximately 15,000 objects of 15 categories, that are scanned from real-world indoor scenes with cluttered backgrounds. 
We test our model on three distinct variants of the dataset, namely `OBJ-BG', `OBJ-ONLY', and `PB-T50-RS'. The results are presented in Tab.~\ref{tab:shapeClsScanObjectNNGfinetune}. On the hardest variant `PB-T50-RS', our model achieves 83.79\% accuracy, surpassing Transformer~\cite{yu2022pointBERT} and Transformer-OcCo~\cite{yu2022pointBERT} and performing slightly inferior to Point-MAE~\cite{pang2022masked}.

\begin{table}[!t]
\caption{Object classification on the real-world dataset, ScanObjectNN. We evaluate out method on three variants, `OBJ-BG', `OBJ-ONLY' and `PB-T50-RS'. Accuracy (\%) for each variant is reported.}
\centering
\begin{tabular}{c|ccc} 
\toprule
Methods             & OBJ-BG & OBJ-ONLY & PB-T50-RS \\ \hline
PointNet~\cite{qi2017pointnet}            & 73.3   & 79.2     & 68.0      \\ 
SpiderCNN~\cite{xu2018spidercnn}           & 77.1   & 79.5     & 73.7      \\ 
PointNet++~\cite{qi2017pointnet++}          & 82.3   & 84.3     & 77.9      \\ 
DGCNN~\cite{wang2019DGCNN}              & 82.8   & 86.2     & 78.1      \\ 
PointCNN~\cite{li2018pointcnn}            & 86.1   & 85.5     & 78.5      \\ 
BGA-DGCNN~\cite{uy2019revisiting}           & -      & -        & 79.7      \\ 
BGA-PN++~\cite{uy2019revisiting}           & -      & -        & 80.2      \\ 
GBNet~\cite{qiu2021geometric}              & -      & -        & 80.5      \\ 
PRANet~\cite{cheng2021net}             & -      & -        & 81.0      \\ \hline
Transformer~\cite{yu2022pointBERT}         & 79.86  & 80.55    & 77.24     \\ 
Transformer-OcCo~\cite{yu2022pointBERT}    & 84.85  & 85.54    & 78.79     \\ 
Point-BERT~\cite{yu2022pointBERT}          & 87.43  & 88.12    & 83.07     \\
Point-MAE~\cite{pang2022masked}           & 90.02  & 88.29    & 85.18     \\ 
Point-MAE (Rep.) & 89.67       & 88.98         & 83.83     \\ 
Ours                & 88.99       & 87.95         & 83.79    \\ 
\bottomrule
\end{tabular}
\label{tab:shapeClsScanObjectNNGfinetune}
\end{table}

\subsection{Part Segmentation} \label{sec:PartSeg}
\par Task of object part segmentation aims to predict a more fine-gained class label for every instance. We conduct part segmentation on ShapeNetPart~\cite{yi2016scalable}, which comprises 16,872 samples shared by 16 categories, divided into 13,998 training and 2,874 testing data. As illustrated in Tab.~\ref{tab:partSeg}, our method achieves competitive results compared to the state-of-the-art methods. Notably, our method outperforms other methods in six categories, namely airplane, cap, knife, laptop, mug and pistol.

\begin{table*}[!t]
\caption{Part segmentation results on ShapeNetPart. We report the mean IoU across all part categories $mIoU_C(\%)$ and the mean IoU across all instances $mIoU_I(\%)$, as well as the IoU(\%) for categories.}
\centering
% \resizebox{18cm}{14mm}{ % width=0.95\linewidth
% \footnotesize
\tabcolsep=1mm
\begin{tabular}{c|cccccccccccccccccc}
\toprule 
\multirow{2}{*}{Methods}    & \multirow{2}{*}{$mIoU_c$}   & \multirow{2}{*}{$mIoU_i$}   & aeroplane     & bag  & cap           & car   & chair & earphone & guitar & knife \\ & &        & lamp & laptop        & motor & mug           & pistol        & rocket & skateboard & table \\ \hline
\multirow{2}{*}{PointNet~\cite{qi2017pointnet}}     & \multirow{2}{*}{80.39}          & \multirow{2}{*}{83.7}           & 83.4          & 78.7         & 82.5         & 74.9         & 89.6           & 73.0               & 91.5            & 85.9 \\ & &          & 80.8          & 95.3            & 65.2           & 93           & 81.2            & 57.9            & 72.8                & 80.6           \\  \hline
\multirow{2}{*}{PointNet++~\cite{qi2017pointnet++}}  & \multirow{2}{*}{81.85}          & \multirow{2}{*}{85.1}           & 82.4          & 79.0           & 87.7         & 77.3         & 90.8           & 71.8               & 91.0              & 85.9  \\ & &          & 83.7          & 95.3            & 71.6           & 94.1         & 81.3            & 58.7            & 76.4                & 82.6           \\ \hline
\multirow{2}{*}{DGCNN~\cite{wang2019DGCNN} }   & \multirow{2}{*}{82.33}          & \multirow{2}{*}{85.2}           & 84.0          & 83.4         & 86.7         & 77.8         & 90.6           & 74.7               & 91.2            & 87.5  \\ & &          & 82.8          & 95.7            & 66.3           & 94.9         & 81.1            & 63.5            & 74.5                & 82.6           \\ \hline
\multirow{2}{*}{Transformer~\cite{yu2022pointBERT}}  & \multirow{2}{*}{83.42}          & \multirow{2}{*}{85.1}           & 82.9          & 85.4         & 87.7         & 78.8         & 90.5           & 80.8               & 91.1            & 87.7    \\ & &        & 85.3          & 95.6            & 73.9           & 94.9         & 83.5            & 61.2            & 74.9                & 80.6           \\ \hline 
\multirow{2}{*}{Transformer-OcCo~\cite{yu2022pointBERT}}   & \multirow{2}{*}{83.42}         & \multirow{2}{*}{85.1}           & 83.3          & 85.2         & 88.3         & 79.9         & 90.7           & 74.1               & 91.9            & 87.6    \\ & &        & 84.7          & 95.4            & 75.5           & 94.4         & 84.1            & 63.1            & 75.7                & 80.8           \\ \hline
\multirow{2}{*}{Point-BERT~\cite{yu2022pointBERT}}   & \multirow{2}{*}{84.11}          & \multirow{2}{*}{85.6}           & 84.3          & 84.8         & 88.0         & 79.8         & 91.0           & 81.7               & 91.6            & 87.9   \\ & &         & 85.2          & 95.6            & 75.6           & 94.7         & 84.3            & 63.4            & 76.3                & 81.5           \\ \hline
\multirow{2}{*}{Point-MAE~\cite{pang2022masked}}  & \multirow{2}{*}{84.19}          & \multirow{2}{*}{\textbf{86.1}}           & 84.3          & \textbf{85.0}         & 88.3         & \textbf{80.5}         & 91.3           & \textbf{78.5}               & \textbf{92.1}            & 87.4   \\ & &         & \textbf{86.1}          & 96.1            & 75.2           & 94.6         & 84.7            & \textbf{63.5}            & \textbf{77.1}                & \textbf{82.4}           \\ \hline
\multirow{2}{*}{Ours}   & \multirow{2}{*}{\textbf{84.2}}  & \multirow{2}{*}{86.0}           & \textbf{84.8} & 84.0         & \textbf{89.2} & 80.2         & \textbf{91.3}           & 77.8               & 92.0            & \textbf{88.3}    \\ & &    & 85.9          & \textbf{96.1}   & \textbf{75.5}           & \textbf{95.0} & \textbf{85.1}   & 62.0            & 75.2                & 82.0           \\ 
\bottomrule
\end{tabular}%}
\label{tab:partSeg}
\end{table*}

% 为了更清晰的展示部分分割的效果，我们选择多个类别分别可视化部分分割的结果，并且与PointMAE部分分割的结果进行对比。 在可视化部分分割图像过程中，我们发现给定的ground truth存在不准确的情况，但是PointMAE以及我们的模型实现了准确分割，在图9中给出可视化的效果图。
% 图像选一下 5个种类就行
\par
\begin{figure*}[!t]
    \centering
    \includegraphics[width=1.0\linewidth]{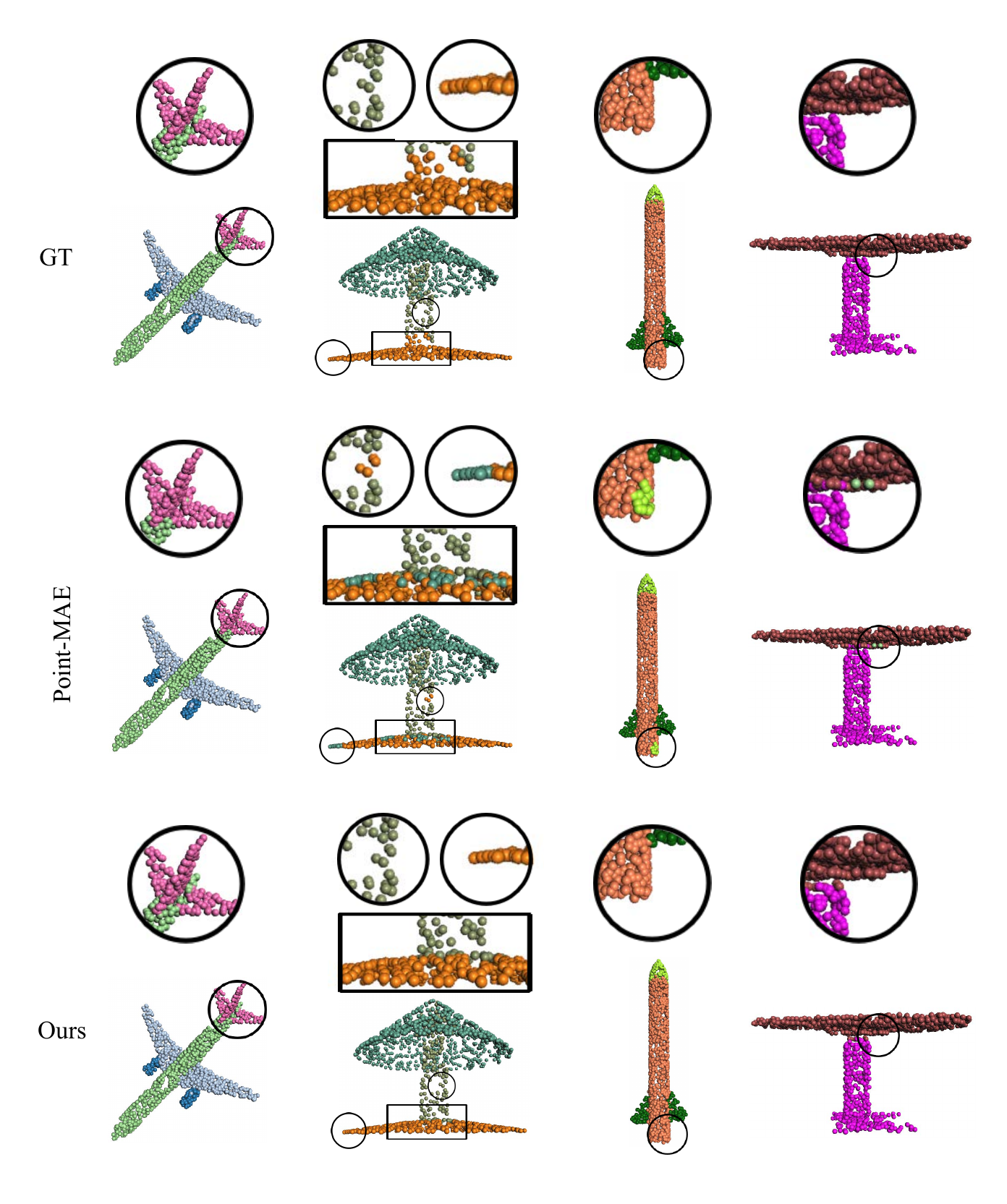}
    \caption{Visualization of the part segmentation results on the ShapeNetPart test set. We choose multiply categories for visualizing the effectiveness of Point-MAE and our method with part segmentation.}
	  \label{fig:multiClass_partSeg_GT}
\end{figure*}

 Subsequently, we visualize the results of both Point-MAE~\cite{pang2022masked} and our method of multiple categories and exhibit them in Fig.~\ref{fig:multiClass_partSeg_GT}.
As evident from the results, our algorithm achieves more precise part segmentation outcomes at the locations where multiple parts interact, owing to the use of geometric descriptors that enable robust extraction of local geometric information.

% \subsubsection{Part Segmentation figure, maskSurf figure 9}
% airplane:   10(lydia/author): 
% airplane: 0(94.17/92.36)   10(93.82/87.04)   13(80.29/76.74)  17(89.41/86.85)  21(96.31/93.95)   24(83.92/77.51)  33(90/83.45)   82(91.81/89)  91(77.09/64.39) 
% 96, 99, 107, 110, 120(*) 125*  133*  %  175
% 备注 airplane 190,238 gt飞机机翼存在标注错误,但是在pointmae和我的中纠正了.
% 备注,lamp 46 吊灯, chair 702,标注错误

\subsection{Semantic Segmentation} \label{sec:SemSeg}
\par We evaluate the performance of our pre-trained model in the task of the 3D semantic segmentation using the Stanford large-scale 3D indoor spaces (S3DIS)~\cite{armeni20163d} dataset. S3DIS is scanned from 272 rooms in 6 indoor areas, with each point annotated into 13 categories. Each area contains 3687, 4440, 1650, 3662, 6852, and 3294 scenes respectively. 
Referring to the semantic segmentation experiments of STRL~\cite{huang2021STRL}, we fine-tune our pre-trained model on one area in Area 1-5, followed by evaluation on Area 6. Furthermore, we supplement the semantic segmentation experiments of Point-MAE~\cite{pang2022masked} based on the pre-trained model released in the official codes. As shown in Tab.~\ref{tab:semSeg}, our model gains a significant improvement of 4.17\%, 4.68\%, 5.49\%, 3.68\%, 3.8\% in accuracy and 11.89\%, 5.93\%, 10.81\%, 5.78\%, 2.95\% in mIoU over the STRL~\cite{huang2021STRL}. Our results demonstrate the capability of our model to extract contextual and semantic information for achieving fine-grained segmentation results.

\begin{table}[!t]
\caption{3D semantic segmentation fine-tuned on S3DIS based on pre-trained model. We report the mean IoU across all class categories $mIoU(\%)$ and the classification accuracy Acc (\%).}
\centering
\begin{tabular}{cccc}
\toprule
Fine-tuning Area                      & Method       & Acc.    & mIoU  \\ \hline
\multirow{3}{*}{Area 1 (3687 samples)}  & STRL~\cite{huang2021STRL}         & 85.28 & 59.15 \\  
                & Point-MAE~\cite{pang2022masked}         & 89.03 & 71.92      \\  
                                       & Ours         & 89.45 & 70.83      \\ \hline
\multirow{3}{*}{Area 2 (4440 samples)}  & STRL~\cite{huang2021STRL}         & 72.37 & 39.21 \\ 
                & Point-MAE~\cite{pang2022masked}         & 76.72 & 47.13      \\   
                                       & Ours         & 77.05 & 45.14      \\ \hline
\multirow{3}{*}{Area 3 (1650 samples)}  & STRL~\cite{huang2021STRL}         & 79.12 & 51.88 \\ 
                & Point-MAE~\cite{pang2022masked}         & 84.09 & 64.29      \\   
                                       & Ours         & 84.61  & 62.69      \\ \hline
\multirow{3}{*}{Area 4 (3662samples)}   & STRL~\cite{huang2021STRL}         & 73.81 & 39.28 \\ 
                & Point-MAE~\cite{pang2022masked}         & 77.34   & 45.15      \\  
                                       & Ours         & 77.49    & 45.06     \\ \hline
\multirow{3}{*}{Area 5 (6852 samples)}  & STRL~\cite{huang2021STRL}     & 77.28 & 49.53 \\ 
                & Point-MAE~\cite{pang2022masked}         & 80.56 & 51.46      \\ 
                                       & Ours         & 81.08   & 52.48      \\ 
\bottomrule
\end{tabular}
\label{tab:semSeg}
\end{table}

\subsection{Ablation Study} \label{sec:Ablation}
\textbf{Ablation study on values of loss.} In order to discover a more optimal pre-trained model for downstream tasks, we conduct a comparative study on three pre-trained models that differ in their pre-training losses. We validate their efficacy through shape classification, followed by global fine-tuning on ModelNet40 to identify the most effective pre-training loss. During testing, we record both the accuracy and the accuracy with the standard voting method~\cite{liu2019relation}, which can enhance the accuracy of experimental outcomes. Comprehensive details regarding the parameters and results are tabulated in Tab.~\ref{tab:ablationEpochs}.

\begin{table}[!t]
\caption{Object classification based on three pre-trained models with different iterations, on ModelNet40 with global fine-tuning protocol. We save pre-trained models with iterations 100, 200 and 300, then choose shape classification for comparing.}
\centering
\begin{tabular}{cccc}
\toprule
Epochs & Loss & Acc.  & Acc+Vote \\ \hline
100    & 1.45 & 92.46 & 93.19    \\ 
200    & 1.25 & 92.54 & 93.23    \\ 
300    & 1.18 & 92.43 & 93.35   \\
\bottomrule
\end{tabular}
\label{tab:ablationEpochs}
\end{table}

\noindent \textbf{Ablation study on three modules.} In order to ascertain the effectiveness of our improved modules, we conducted ablation experiments with geometric descriptors, adaptive saliency attention, and external attention modules. We pre-trained these modules for 300 epochs and evaluated their performance on shape classification using the ModelNet40 dataset with global fine-tuning protocol. The results of our experiments are presented in Tab.~\ref{tab:ablationModules}.

\begin{table}[!t]
\caption{Object classification based on ModelNet40 with global fine-tuning protocol. We separately incorporate the geometric descriptor, adaptive saliency attention, and external attention module for pre-training and evaluate their effectiveness through shape classification on ModelNet40.}
\centering
\begin{tabular}{ccc}
\toprule
Methods                   & Acc   & Acc+Vote \\ \hline
+Descriptor                & 92.75 & 93.31    \\ 
+Adaptive Saliency Attention & 92.70 & 93.11    \\ 
+External Attention        & 92.46 & 93.23   \\ 
Ours        & 92.43 & 93.35   \\ 
\bottomrule
\end{tabular}
\label{tab:ablationModules}
\end{table}

\section{Conclusion} \label{sec:Conculsion}
\par In this paper, we propose PointGame, a concise and effective method for point cloud self-supervised representation learning. For selectively focusing on salient parts and efficiently extracting significant features, we design the adaptive saliency attention mechanism to obtain tokens. Additionally, we adopt geometric descriptors to extract local geometry information, and conduct descriptor embedding to obtain descriptor tokens. We then merge the tokens of point patches and geometric descriptors, and utilize them as the input for the auto-encoder. 
Unlike previous works with standard transformer blocks with self-attention mechanism, we utilize external attention to lower the computational complexity while maintaining performance, and incorporate correlations between different samples for a more descriptive feature representation.
\par To showcase the versatility and adaptability of our proposed method, PointGame, we conduct several downstream tasks to validate its generalization capabilities, including shape classification, object part segmentation, and semantic segmentation. For shape classification experiments, we fine-tune model with global and local fine-tuning strategies. Despite outperforming most self-supervised learning methods, our method's performance on real-world datasets can still be improved. Therefore, we plan to further enhance the domain adaptation ability to achieve better results on real-world datasets.

% \section*{Acknowledgments}
% This should be a simple paragraph before the References to thank those individuals and institutions who have supported your work on this article.

% \section{References}
\bibliographystyle{IEEEtran}
\bibliography{IEEEfull}

\vspace{11pt}


\vfill

\end{document}